\documentclass[sigconf,nonacm]{acmart}
\AtBeginDocument{%
  }

\copyrightyear{2025}
\acmYear{2025}
\setcopyright{acmlicensed}\acmConference[WWW Companion '25]{Companion Proceedings of the ACM Web Conference 2025}{April 28-May 2, 2025}{Sydney, NSW, Australia}
\acmBooktitle{Companion Proceedings of the ACM Web Conference 2025 (WWW Companion '25), April 28-May 2, 2025, Sydney, NSW, Australia}
\acmDOI{10.1145/3701716.3717647}
\acmISBN{979-8-4007-1331-6/2025/04}

\usepackage{cleveref}
\usepackage{xcolor} 
\usepackage{enumitem}

\usepackage{flushend}
\usepackage{balance}

\begin{document}

\title{Federated Fine-Tuning of Large Language Models:\\ Kahneman-Tversky vs. Direct Preference Optimization}

\author{Fernando Spadea}
\orcid{0009-0006-4278-3666}
\affiliation{%
  \institution{Rensselaer Polytechnic Institute}
  \streetaddress{110 8th St}
  \city{Troy}
  \state{NY}
  \country{USA}
}
\email{spadef@rpi.edu}

\author{Oshani Seneviratne}
\orcid{0000-0001-8518-917X}
\affiliation{%
  \institution{Rensselaer Polytechnic Institute}
  \streetaddress{110 8th St}
  \city{Troy}
  \state{NY}
  \country{USA}
}
\email{senevo@rpi.edu}

\renewcommand{\shortauthors}{Spadea and Seneviratne}
\renewcommand{\shorttitle}{Federated Fine-Tuning of Large Language Models}

\begin{abstract}
  We evaluate Kahneman-Tversky Optimization (KTO) as a fine-tuning method for large language models (LLMs) in federated learning (FL) settings, comparing it against Direct Preference Optimization (DPO). Using Alpaca-7B as the base model, we fine-tune on a realistic dataset under both methods and evaluate performance using MT-Bench-1, Vicuna, and AdvBench benchmarks. Additionally, we introduce a redistributed dataset setup, where only KTO is applicable due to its ability to handle single-response feedback, unlike DPO’s reliance on paired responses. Our results demonstrate that KTO, in both its original (KTOO) and redistributed (KTOR) configurations, consistently outperforms DPO across all benchmarks. In the redistributed setup, KTO further validates its flexibility and resilience by maintaining superior performance in scenarios where DPO cannot be applied. These findings establish KTO as a robust and scalable fine-tuning method for FL, motivating its adoption for privacy-preserving, decentralized, and heterogeneous environments.
\end{abstract}



\maketitle

\section{Introduction}

Federated learning (FL) is a rapidly growing field focused on training AI models in a decentralized manner~\cite{mcmahan2017communication}. By enabling a central server to share a model with several clients for local training, FL allows these models to learn from private data without requiring data centralization, thus preserving privacy and complying with regulatory frameworks such as GDPR. Additionally, FL addresses the scalability and efficiency challenges associated with training LLMs and other foundation models, which are characterized by their massive scale and broad applicability~\cite{bommasani2021opportunities}. 

\subsection{The Need for Fine-Tuning in Federated Learning}
Fine-tuning is a crucial stage in the training of LLMs, enabling them to adapt to specific tasks, domains, or datasets. Unlike pre-training, which focuses on learning general representations from large, diverse datasets, fine-tuning refines a model's parameters for narrower applications such as sentiment analysis, instruction following, or domain-specific question answering \cite{howard2018universal}. In the context of FL, fine-tuning is essential to deploy LLMs in privacy-sensitive and distributed environments, as it allows training on decentralized datasets while preserving privacy and ensuring compliance with data protection regulations. Moreover, fine-tuning in FL enables model personalization for diverse users or organizations and addresses the challenges of data heterogeneity and resource constraints in real-world applications~\cite{smith2017federated}.

\subsection{Challenges in Fine-Tuning for Federated Learning}
Fine-tuning LLMs in FL settings introduces unique challenges:
\begin{itemize}[noitemsep, leftmargin=0pt]
    \item \textbf{Data Availability and Heterogeneity:} Data is often non-independent and identically distributed data (non-IID) across clients, leading to inconsistencies in the training process.
    \item \textbf{Resource Constraints:} Clients in FL often have limited computational power, making the resource-intensive nature of LLM fine-tuning a significant hurdle~\cite{kairouz2021advances}.
    \item \textbf{Privacy Concerns:} Sensitive client data cannot be centralized, restricting the application of conventional fine-tuning techniques.
\end{itemize}

\subsection{Fine-Tuning Methods}
This paper explores two fine-tuning methods for LLMs in FL settings: DPO and KTO. These methods differ significantly in their data requirements and applicability:
\begin{itemize}[noitemsep, leftmargin=0pt]
    \item \textbf{Direct Preference Optimization (DPO):} DPO relies on paired responses for each input (e.g., one labeled "good" and the other "bad") and optimizes the model to prefer the better response~\cite{rafailov2024direct}. While effective, DPO is data-intensive and less resilient to heterogeneity, making it less suitable for FL scenarios where data availability is limited and unevenly distributed.
    \item \textbf{Kahneman-Tversky Optimization (KTO):} KTO is a simpler, more flexible fine-tuning method that requires only a single response per prompt, labeled as "good" or "bad"~\cite{ethayarajh2024kto}. This reduced complexity makes KTO more adaptable to FL environments, particularly when data is non-IID or sparsely distributed across clients.
\end{itemize}

\subsection{Our Contributions}
In this work, we focus on fine-tuning LLMs in FL settings, with an emphasis on the simplicity and resilience of KTO. Specifically, we:
\begin{enumerate}
    \item Evaluate KTO against DPO across various FL aggregation algorithms (e.g., FedAvg~\cite{mcmahan2017communication}, FedProx~\cite{li2020federated}) to assess its performance in decentralized settings.
    \item Demonstrate that KTO effectively addresses data heterogeneity and limited availability, outperforming DPO in most scenarios.
    \item Highlight the advantages of KTO in reducing resource requirements and improving privacy preservation in FL environments.
\end{enumerate}

The remainder of this paper is organized as follows: In Section~\ref{sec:related}, we review related work on fine-tuning LLMs in FL settings. Section~\ref{sec:design} describes the experimental design and methodology for comparing KTO and DPO. Section~\ref{sec:results} presents our results and discusses the implications of KTO’s performance. We discuss potential potential research directions in section~\ref{sec:discussion}. Finally, Section~\ref{sec:conclusion} summarizes our findings and outlines directions for future research.

\section{Related Work}
\label{sec:related}

FL has gained significant attention as a privacy-preserving paradigm for training AI models across decentralized data sources. Notably, Ye et al. \cite{ye2024openfedllm, ye2024fedllm} have designed realistic benchmarks for FL tailored to LLMs. These benchmarks incorporate a diverse range of aggregation methods, including FedAvg \cite{mcmahan2017communication}, FedProx \cite{li2020federated}, SCAFFOLD \cite{karimireddy2020scaffold}, FedAvgM \cite{hsu2019measuring}, FedYogi \cite{reddiadaptive}, FedAdagrad \cite{reddiadaptive}, and FedAdam \cite{reddiadaptive}, each of which offers unique approaches to combining model updates from distributed clients at a central server.

A key strength of these benchmarks is their use of real-world data from Chatbot-arena Conversations \cite{zheng2023judging}. This dataset consists of authentic human-chatbot interactions, organized by User ID to simulate clients in a federated learning setting, thereby addressing the heterogeneity and sparsity challenges typical in FL scenarios. 
In particular, Ye et al.'s benchmarks evaluate performance across three critical metrics: 
\begin{enumerate}
    \item \textbf{MT-Bench-1:} A benchmark for assessing 1-turn conversational abilities \cite{zheng2023judging}.
    \item \textbf{Vicuna:} A benchmark designed to evaluate instruction-following capabilities \cite{chiang2023vicuna}.
    \item \textbf{AdvBench:} A benchmark focused on AI safety, measuring a model’s robustness to adversarial prompts \cite{zou2023universal}.
\end{enumerate}

Their results demonstrate decent performance for DPO across these benchmarks. However, DPO's reliance on paired responses and its susceptibility to data heterogeneity limit its applicability in real-world FL scenarios. In our work, we build upon these benchmarks and demonstrate that KTO, a simpler and more flexible fine-tuning method, outperforms DPO in terms of both effectiveness and adaptability. 



\section{Evaluation Methodology}
\label{sec:design}

\subsection{Model and Fine-Tuning Setup}
The base model used in our experiments is Alpaca-7B, an instruction-tuned model built on LLAMA-7B \cite{alpaca}. We fine-tune the model using LoRA \cite{hu2021lora}, a parameter-efficient fine-tuning method, without quantization to ensure consistency across all the benchmarks we were testing.
The training process largely mirrors that of the original paper, except that we employ KTO instead of DPO. 

\begin{figure}
  \centering
  \includegraphics[width=0.6\columnwidth]{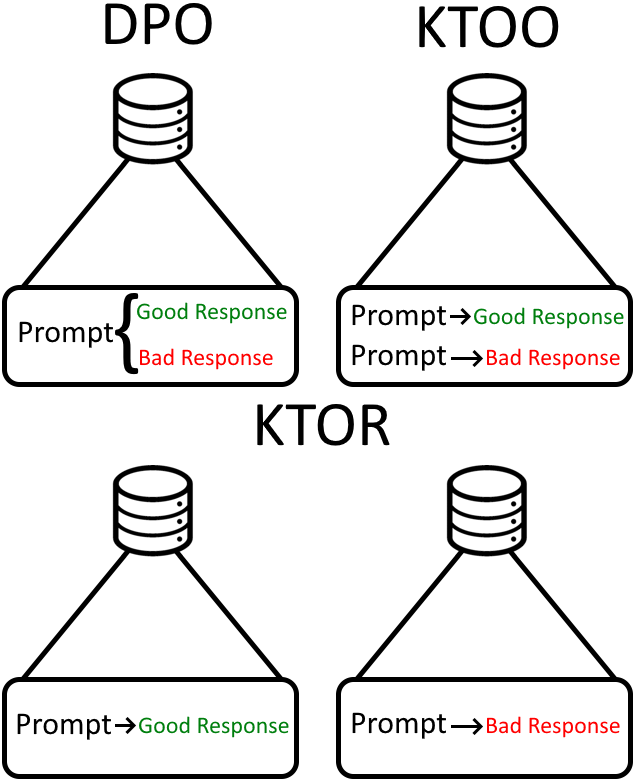}
  \caption{Difference in data design between DPO, KTOO, and KTOR.}
  \label{fig:dpokto}  
\end{figure}

To adapt the data for KTO, each DPO data point, which consists of paired "good" and "bad" responses, is split into two separate data points with individual labels indicating the response quality. The transformation process is illustrated in \Cref{fig:dpokto}. This adaptation allows for a direct comparison between KTO and DPO under identical conditions.

\subsection{Benchmark Design}
To evaluate the performance of KTO, we conduct two rounds of experiments:
\begin{enumerate}
    \item \textbf{Original Data Allocation (KTOO):} Data points remain associated with their original clients, providing a direct comparison between DPO and KTO under the same client data distribution.
    \item \textbf{KTO Redistributed (KTOR):} Data points are randomly reassigned between clients while maintaining the same number of data points per client. This is performed pseudorandomly with a fixed seed (2023) to ensure consistency across experiments. This setup tests KTO’s robustness when the "good" and "bad" responses for prompts are distributed across different clients, a scenario where DPO cannot be applied.
\end{enumerate}

\subsection{Aggregation Methods and Benchmarks}
We evaluate the fine-tuned models using the same aggregation methods as the original study, including:
\begin{itemize}[noitemsep, leftmargin=0pt]
    \item \textbf{Aggregation Methods:} FedAvg \cite{mcmahan2017communication}, FedProx \cite{li2020federated}, SCAFFOLD \cite{karimireddy2020scaffold}, FedAvgM \cite{hsu2019measuring}, FedYogi \cite{reddiadaptive}, FedAdagrad \cite{reddiadaptive}, and FedAdam \cite{reddiadaptive}.
    \item \textbf{Evaluation Benchmarks:} MT-Bench-1 \cite{zheng2023judging} for one-turn conversational ability, Vicuna \cite{chiang2023vicuna} for instruction-following tasks, and AdvBench \cite{zou2023universal} for AI safety.
\end{itemize}

While AdvBench employs a rule-based keyword search to evaluate model outputs, MT-Bench-1 and Vicuna require human-like judgment to assign scores out of 10. Traditionally, this is done using GPT-4, but we use JudgeLM-13B \cite{zhu2023judgelm}, an open-source alternative. JudgeLM-13B evaluates model outputs collectively across DPO, KTOO, and KTOR fine-tuning methods, ensuring that scores reflect relative performance differences while accounting for any systematic bias in the judge model.

\subsection{Base Model Comparisons}
For comparison purposes, we also evaluate the base Alpaca-7B model using FedAvg outputs from DPO and KTO. While these base model results provide a baseline for improvement, we note that the MT-Bench-1 and Vicuna scores for the base model may not be fully reliable due to differences in judgment criteria between JudgeLM-13B and GPT-4.

In total, we evaluate three fine-tuning methods (DPO, KTOO, KTOR) across seven aggregation methods, resulting in 42 benchmark results. Additionally, three benchmark results are provided for the base model. This comprehensive setup allows for a thorough comparison of KTO’s performance against DPO and provides insights into its robustness and efficiency in FL scenarios.

\section{Results}
\label{sec:results}

\begin{table*}[t]
    \centering
    \begin{tabular}{l|l|l|l||l|l|l||l|l|l}
     \hline
     \textbf{Aggr. Method} & \multicolumn{3}{l||}{\textbf{MT-Bench-1 (/10)}} & \multicolumn{3}{l||}{\textbf{Vicuna (/10)}} & \multicolumn{3}{l}{\textbf{AdvBench (/100)}} \\ [0.5ex] 
     \hline\hline
     Base & \multicolumn{3}{c||}{7.51} & \multicolumn{3}{c||}{7.51} & \multicolumn{3}{c}{9.62} \\ 
     \hline
     {} & \textbf{DPO} & \textbf{KTOO} & \textbf{KTOR} & \textbf{DPO} & \textbf{KTOO} & \textbf{KTOR} & \textbf{DPO} & \textbf{KTOO} & \textbf{KTOR} \\
     \hline
     FedAvg & 7.84 & 8.14 \textcolor{green}{$\uparrow$} & 8.11 & 8.03 & 8.51 \textcolor{green}{$\uparrow$} & 8.40 & 12.50 & 15.77 \textcolor{green}{$\uparrow$} & 12.69 \\
     FedProx & 7.73 & 8.44 \textcolor{green}{$\uparrow$} & 8.01 & 7.73 & 8.39 \textcolor{green}{$\uparrow$} & 8.23 & 13.08 & 16.15 \textcolor{green}{$\uparrow$} & 15.58 \\
     SCAFFOLD & 8.01 & 8.17 \textcolor{green}{$\uparrow$} & 7.83 & 7.91 & 8.34 \textcolor{green}{$\uparrow$} & 8.21 & 14.23 & 14.62  & 17.50 \textcolor{green}{$\uparrow$} \\
     FedAvgM & 7.16 & 7.54 & 7.56 \textcolor{green}{$\uparrow$} & 7.84 & 7.99  & 8.37 \textcolor{green}{$\uparrow$} & 8.65 & 12.31  & 14.81 \textcolor{green}{$\uparrow$} \\
     FedYogi & 8.75 & 8.98 & 9.03 \textcolor{green}{$\uparrow$}& 7.65 & 8.21 \textcolor{green}{$\uparrow$} & 8.13 & 11.35 & 12.88  & 17.12 \textcolor{green}{$\uparrow$}\\
     FedAdagrad & 8.49 & 8.84 \textcolor{green}{$\uparrow$} & 8.78 & 7.96 & 8.32  & 8.34 \textcolor{green}{$\uparrow$} & 11.54 & 12.88 \textcolor{green}{$\uparrow$} & 11.92 \\
     FedAdam & 8.20 & 8.64 \textcolor{green}{$\uparrow$} & 8.43 & 7.89 & 8.55 \textcolor{green}{$\uparrow$} & 8.47 & 11.35 & 12.69 & 13.46 \textcolor{green}{$\uparrow$}\\ [1ex] 
     \hline
    \end{tabular}
    \caption{Benchmark Results: KTOR and KTOO represent the models trained with KTO that did and did not have the data redistributed prior to training, respectively. MT-Bench-1 and Vicuna are average scores out of 10, while AdvBench is out of 100. Green arrows indicate which fine-tuning method performed best for each benchmark.}
    \label{table:results}
\end{table*}

The benchmark results, shown in \Cref{table:results}, demonstrate that KTO Original (KTOO) consistently outperforms DPO across all evaluation metrics, including 1-turn conversational ability (MT-Bench-1), instruction following (Vicuna), and AI safety (AdvBench). This highlights its potential for improving LLM fine-tuning under practical constraints like data heterogeneity and privacy.
Additionally, KTO Redistributed (KTOR) outperforms DPO in all but one instance (MT-Bench-1 with SCAFFOLD). Interestingly, KTOR even surpasses KTO in certain scenarios, highlighting its robustness in handling redistributed and heterogeneous data. This improvement is likely due to the redistribution of good and bad responses for prompts across clients, which alters the influence of individual prompts on the aggregated model depending on the chosen aggregation method. While this redistribution may reduce training quality for a single client, it has the potential to enhance the overall quality of federated learning. Our results indicate that the impact of this phenomenon varies across aggregation methods and benchmarks, but KTOR consistently outperformed KTOO with FedAvgM across all three benchmarks. FedAvgM’s design, which better handles client distribution differences, explains its superior performance with KTOR compared to FedAvg, which consistently favored KTOO.

One caveat of our results is the reliance on JudgeLM-13B as the evaluation model, which tends to produce higher absolute scores compared to GPT-4. As noted in Section~\ref{sec:design}, JudgeLM-13B graded our three fine-tuning methods (DPO, KTO, and KTOR) collectively for each benchmark to ensure consistent scores relative to each other. While the raw scores may differ from GPT-4-based evaluations in other studies, the relative performance trends within our experiments remain valid. Thus, our findings robustly demonstrate that KTOO and KTOR significantly outperform DPO in federated learning contexts.

These results reinforce the potential of KTO as a flexible and scalable fine-tuning method for LLMs in federated environments.



\section{Discussion}
\label{sec:discussion}

Our results demonstrate the effectiveness of KTO as a robust fine-tuning method for FL. However, there are several critical areas for future research that could enhance the understanding and adoption of KTO in federated learning. Addressing these aspects would further solidify KTO’s position as a flexible, scalable, and efficient fine-tuning method for LLMs in FL scenarios, enabling real-world applications across diverse domains. Key areas for further exploration include:

\begin{itemize}[noitemsep, leftmargin=0pt]
    \item \textbf{Impact of Quantization}

To ensure consistency, all training in this study was conducted without quantization, diverging from Ye et al.'s original study, which utilized 8-bit quantization. This approach allowed us to attribute performance differences directly to the fine-tuning methods rather than to quantization artifacts. Nonetheless, exploring quantization, such as 8-bit or lower precision, in future studies could provide insights into KTO’s performance under resource-constrained settings. Quantization could enhance the efficiency of FL deployments while retaining KTO’s demonstrated strengths.

    \item \textbf{Evaluation Model Choice}

JudgeLM-13B was chosen as the evaluation model for its internal consistency across all benchmarks. While effective for this study, employing GPT-4, a widely recognized standard, would facilitate direct comparisons with other studies. Additionally, incorporating evaluations from multiple judges could further validate the robustness of KTO and KTOR across diverse assessment frameworks, providing a more comprehensive evaluation of their effectiveness.

    \item \textbf{Dataset Design and Redistribution}

Our evaluation relied on a redistributed version of the DPO dataset to create the KTOR setup. This demonstrated KTO’s resilience in handling heterogeneous client data. However, future work could focus on developing a dedicated dataset tailored to KTO’s single-response feedback framework. A dataset designed explicitly for KTO would minimize biases introduced by adapting datasets created for DPO and provide a more realistic foundation for testing KTO’s capabilities.

    \item \textbf{Diverse Metrics}

To further evaluate the versatility of KTO, future work should incorporate additional benchmarks, such as MMLU \cite{hendrycks2020measuring}. MMLU provides a comprehensive assessment of a model’s knowledge across a wide range of tasks. By extending the analysis beyond conversational ability, instruction-following, and safety tasks, these benchmarks would offer a broader understanding of KTO’s capabilities.

    \item \textbf{Resource Constraints and Efficiency}

Federated learning scenarios often operate under significant resource constraints, making computational efficiency a critical factor. While this study focused on benchmarking performance, future work should measure and compare resource utilization for DPO and KTO. Metrics such as memory usage, communication overhead, and convergence rates could provide practical insights into KTO’s scalability and its viability in real-world FL deployments.

\end{itemize}

\section{Conclusion}
\label{sec:conclusion}

This work demonstrates the effectiveness of KTO as a fine-tuning method for LLMs in FL environments. Our results show that KTO consistently outperforms DPO across various benchmarks, even when evaluated under the same datasets and client distributions. Moreover, KTO’s adaptability becomes particularly evident in scenarios involving redistributed datasets (KTOR), where DPO is no longer applicable. This highlights KTO’s resilience to heterogeneous data, a critical challenge in real-world FL scenarios.

KTO’s flexibility and robust performance make it a compelling choice for federated fine-tuning. Unlike DPO, which relies on paired "good" and "bad" responses, KTO operates effectively with single-response feedback, expanding its applicability to a wider range of FL applications. Furthermore, KTO achieves superior results in key tasks such as conversational ability, instruction-following, and AI safety, emphasizing its potential to enhance LLM fine-tuning in privacy-preserving, decentralized settings.



\section{Resource Contributions}

Our research artifacts (code, documentation, and graphs) are shared under the Apache 2.0 license. We maintain an open-source GitHub repository for all our artifacts: \url{https://github.com/brains-group/OpenFedLLM/tree/KTO}.

\balance

\bibliographystyle{ACM-Reference-Format}
\bibliography{references.bib}

\end{document}